\DocumentMetadata{
	lang		= eng-US, 	
	pdfstandard	= A-3u,		
	pdfversion	= 1.7, 		
}

\documentclass[nohead,colorlinks,upint,subscriptcorrection,varvw,hyphenate,balance,greek,russian,vietnamese,german]{asmeconf} 

\usepackage{makecell, multirow}
\usepackage{algorithm}
\usepackage{algorithmic}
\usepackage{amsmath}

\usepackage[dvipsnames,table]{xcolor}
\definecolor{mydarkgreen}{RGB}{0,120,0}
\definecolor{mydarkred}{RGB}{170,0,0}
\usepackage{pifont}
\newcommand{\CheckmarkBold}{\ding{51}} 

\usepackage{hyperref}
\usepackage{hyperxmp}

\usepackage{makecell}
\usepackage{tabularx}

\usepackage{kotex} 

\usepackage{xcolor} 

\hypersetup{%
	pdfauthor={},
	pdftitle={},
	pdfkeywords={},
	pdfsubject = {},
}

\allowdisplaybreaks 

\begin{document}



%

\title{Enhancing Creativity in 3D Generative Design via a TRIZ-Inspired Text-to-CAD Framework} 
 
%
%
%

\SetAuthors{%
	Dongeon Lee\affil{1},
	Leekyo Jeong\affil{1},
	Soyoung Yoo\affil{2},
    Sunwoong Yang\affil{3},
	Namwoo Kang\affil{1}\affil{4}\CorrespondingAuthor{nwkang@kaist.ac.kr}
	}

\SetAffiliation{1}{Cho Chun Shik Graduate School of Mobility, KAIST, Daejeon, Republic of Korea}
\SetAffiliation{2}{Samsung Electronics, Suwon, Republic of Korea}
\SetAffiliation{3}{Department of Mechanical Engineering, Hanyang University, Ansan, Republic of Korea}
\SetAffiliation{4}{Narnia Labs, Daejeon, Republic of Korea}



\maketitle





\begin{abstract}
Recent advances in large language models (LLMs) have demonstrated significant potential in supporting engineering design tasks, including computer-aided design (CAD) automation. However, most existing LLM-based 3D CAD generation approaches primarily focus on geometric precision and instruction-following performance, often overlooking the fundamental aspect of creative design exploration. This study presents a TRIZ-inspired text-to-CAD framework that leverages LLMs to generate high-quality, editable CAD models while systematically exploring creative design alternatives. The framework integrates the Theory of Inventive Problem Solving (TRIZ)—embedding deep human insights from extensive patent records—into LLM prompting strategies, enabling autonomous generation of innovative CAD variants that address technical contradictions. Through a comprehensive three-stage pipeline of design generation, enhancement, and optimization, the framework produces structurally diverse CAD models from well-crafted prompts. The present study implements and evaluates the first two stages, while positioning the design optimization stage as future work. A product design case study (chair) demonstrates that the TRIZ-inspired text-to-CAD framework generates multiple creative design alternatives by systematically applying TRIZ inventive principles such as segmentation, anti-weight, dynamics, and composite materials, achieving 4.0-14.7\% mass reduction across all enhanced designs while maintaining structural integrity. The key findings suggest that integrating systematic innovation methodologies with LLM-based 3D CAD generation bridges the gap between precision-focused synthesis and creativity-focused exploration, advancing toward autonomous design systems where AI makes design decisions independently, supporting human decision-making in human-AI collaborative design for engineering applications.
\end{abstract}


\section{Introduction}
\label{sec:introduction}
Large language models (LLMs) are increasingly being integrated into engineering design and manufacturing workflows, supporting tasks from conceptual to detailed design \cite{picard2025concept}, with applications including text-to-design, design space exploration, design for manufacturing (DfM), performance prediction, and inverse design \cite{makatura2023can}. Recent studies have shown that LLMs can directly generate command code for computer-aided design (CAD), computer-aided manufacturing (CAM), and computer-aided engineering (CAE) systems from natural language input, significantly lowering barriers to design and analysis automation \cite{makatura2023can, guo2026large}. In parallel, increasing attention has been given to LLM-based text-to-CAD frameworks across various industrial domains, where natural language specifications are directly translated into executable CAD code or parametric representations \cite{zhang2025large}.

However, most existing LLM-based text-to-CAD approaches emphasize precise and instruction-following CAD generation with high geometric fidelity \cite{khan2024text2cad, badagabettu2024query2cad, sun2025large}. Recent vision language model (VLM)-based approaches have incorporated multimodal inputs into CAD code generation, yet continue to prioritize syntactic correctness and geometric accuracy \cite{alrashedy2024generating, li2025llm4cad, doris2026cad}. While precision is essential for downstream manufacturing, such precision-focused approaches often overlook a fundamental aspect of engineering design: the systematic exploration of creative design alternatives. Recent studies have begun to explore prompt-based strategies for design space exploration in generative AI-driven product design, demonstrating that prompt formulation can significantly influence the diversity and creativity of generated design concepts \cite{ma2023conceptual,chong2025prompting,ma2025large}. More recently, prompting strategies in generative AI systems have increasingly been viewed as optimizable control variables within LLM-driven workflows, enabling systematic refinement to improve performance and controllability across downstream tasks \cite{fang2025comprehensive}. Notably, domain-specific prompt structuring and formatting have emerged as critical assets across diverse engineering disciplines, with tailored prompt architectures becoming fundamental enablers for controllable and high-quality generative outputs. Nevertheless, most of these prompt-based and generative approaches have largely remained at the text-to-image generation level, producing visual design concepts rather than executable, parametric CAD representations. As a result, the proposed methodologies have limited applicability to downstream engineering design workflows, where editability, performance validation, and manufacturability are essential. This observation raises a central research question: \textit{How can we effectively leverage LLMs as autonomous design agents to generate high-quality, parametrically editable CAD models, while supporting human-AI collaborative exploration of the creative design space?}

To address this gap, we propose integrating the Theory of Inventive Problem Solving (TRIZ) into LLM-based CAD code generation workflows. TRIZ is a systematic methodology for innovation derived from the analysis of millions of patents, providing structured tools—such as 40 inventive principles and a contradiction matrix—that guide designers toward creative solutions by resolving technical trade-offs \cite{altshuller1984creativity}. Despite its demonstrated potential in conceptual design and ideation, TRIZ has historically faced limitations in systematic validation, industrial diffusion, and computational integration \cite{ilevbare2013review, chechurin2016understanding}. These challenges suggest an opportunity to embed structured innovation principles into modern AI-driven design workflows. While multiple strategies exist for leveraging LLMs in engineering applications, including prompt engineering, retrieval-augmented generation (RAG), and fine-tuning \cite{zhang2025large, chong2025prompting, alammar2024hands}, this study focuses on prompt engineering as a practical, accessible, and user-friendly approach. Specifically, prompt engineering enables the incorporation of TRIZ-informed contextual guidance into widely available closed-source LLMs, such as ChatGPT, Claude, and Gemini, without requiring additional infrastructure and model customization.

Our framework operates through three integrated stages: (1) design generation using domain-specific prompt engineering to create a baseline CAD model, (2) design enhancement through TRIZ-inspired analysis to generate creative alternatives, and (3) design optimization integrating computational analysis—such as CAE simulation and surrogate models—for performance refinement. By embedding TRIZ context into LLM prompts, this approach transforms LLMs from passive instruction-following code generators into autonomous design agents capable of systematically addressing design contradictions and generating diverse, meaningful alternatives. This agent-oriented perspective aligns with a vision in which AI systems in engineering design are increasingly envisioned as actively transitioning from passive automation tools toward autonomous collaborative design partners that inspire human creativity and support systematic design space exploration through human-AI teaming \cite{goucher2026future}. Recent studies have highlighted the emergence of agentic multi-agent AI systems in engineering design, demonstrating that structured orchestration of LLM-based agents can enhance conceptual design processes and move toward more autonomous, end-to-end design exploration \cite{jiang2025intelligent, liu2026idesigngpt}. Building upon advances in generative AI-driven design and optimization frameworks \cite{oh2019deep, yoo2021integrating, jang2022generative}, this work extends emerging multi-agent design systems \cite{jadhav2024large, elrefaie2025ai} toward autonomous design exploration in early-stage conceptual design.

The main contributions and novelties of this work are summarized as follows:

\begin{itemize}
    \item A TRIZ-inspired text-to-CAD framework is proposed to systematically explore LLMs' creative design generation potential through structured prompt engineering, extending recent generative design frameworks \cite{kang2025generative} toward creative exploration in early-stage conceptual design.
    \item A representative product design case study demonstrates that LLMs can generate multiple diverse and creative CAD models from a single baseline while preserving parametric editability and geometric validity, using widely available closed-source LLMs without fine-tuning.
    \item Practical insights are derived illustrating how knowledge-guided prompt engineering can effectively bridge precision-focused 3D CAD generation and systematic creative design exploration, thereby advancing autonomous design systems leveraging LLM capabilities.
\end{itemize}

The paper is organized as follows: Section~\ref{sec:background} reviews LLM-based CAD code generation and TRIZ fundamentals, including the contradiction matrix and prior integrations. Section~\ref{sec:methodology} presents the TRIZ-inspired text-to-CAD framework and its structured prompt engineering strategy. Section~\ref{sec:results_and_discussion} describes the experimental setup and case study results, followed by discussion of practical implications and limitations. Section~\ref{sec:conclusions_and_future_work} summarizes key findings and future research directions.


\section{Background}
\label{sec:background}
This section outlines the key concepts and technical background required to contextualize the proposed TRIZ-inspired text-to-CAD framework. In particular, it reviews LLM-based CAD code generation and TRIZ concepts that form the foundation of the subsequent design framework.

\subsection{LLM-based CAD code generation}
\label{subsec:llm_cad}
Recent studies on LLM-based CAD code generation have shown that natural language queries can be translated into parametric CAD code, enabling the automated generation of editable 3D models \cite{zhang2025large}. Existing approaches include multimodal input processing, iterative code refinement, and model fine-tuning on CAD-specific datasets. In this context, prompt engineering has emerged as a practical and effective mechanism for specifying CAD generation tasks without model fine-tuning \cite{zhang2025large, alammar2024hands}. By structuring input instructions to encode design intent, geometric constraints, and parametric specifications, prompt engineering enables functional CAD code generation in zero-shot settings, providing a flexible interface for guiding closed-source LLM-based text-to-CAD workflows.

Following established best practices in LLM prompting \cite{alammar2024hands}, recent prompt engineering strategies typically structure prompts using clearly defined elements, including (1) role specification to contextualize the model as a domain expert, (2) explicit task definitions with design objectives and constraints, (3) detailed functional and geometric requirements, and (4) well-defined expected output that prescribes executable CAD code workflows. Such structured prompts have been shown to improve syntactic correctness and executability of generated CAD code, while maintaining adherence to design specifications. In addition, iterative prompt refinement—where execution errors are fed back to the model for correction—serves as a practical mechanism for handling code generation failures in text-to-CAD workflows. 

Despite these advances, most existing approaches continue to emphasize precision and instruction-following CAD generation, often at the expense of systematic creative design exploration. These methods excel at geometric accuracy but lack structured frameworks to systematically generate and evaluate innovative alternative design concepts that address conflicting engineering requirements. As a result, LLMs tend to reproduce conventional geometries rather than generating diverse, creative, and inventive design alternatives.

\subsection{TRIZ: Theory of Inventive Problem Solving}
\label{subsec:triz}
TRIZ was originally developed by Genrich Altshuller as a systematic methodology for innovation derived from the analysis of patterns observed in millions of patents \cite{altshuller1984creativity}. The fundamental premise of TRIZ is that technological evolution follows identifiable trends, and that creative solutions to complex engineering problems can be systematically generated by resolving inherent technical contradictions using universal inventive principles \cite{ilevbare2013review, chechurin2016understanding}. TRIZ provides a structured framework for creative problem solving through contradiction analysis and a set of 40 inventive principles, and has been widely applied in engineering design contexts, including product design processes. Recent studies have further demonstrated the applicability of TRIZ-based methodologies in contemporary product design contexts, including conceptual design and additive manufacturing, where structured principle-based guidance enhances creative ideation and reduces design iteration cycles \cite{mazlan2022development, linhao2026triz}.

\subsubsection{Fundamental concepts}
\label{subsubsec:fundamental_concepts}
TRIZ is built upon several core concepts that enable systematic innovation:

\begin{itemize}
    \item \textbf{Technical contradictions}: situations where improving one system attribute leads to the deterioration of another, representing the core design conflicts that TRIZ seeks to resolve.
    \item \textbf{40 inventive principles}: a set of generalized strategies for resolving technical contradictions, distilled from the analysis of millions of patents across diverse engineering domains.
    \item \textbf{Contradiction matrix}: a structured tool that maps pairs of conflicting system parameters to corresponding inventive principles, enabling designers to systematically identify the most promising solution strategies for a given conflict.
    \item \textbf{Ideal final result (IFR)}: the concept of an ideal system that delivers maximum functionality without associated costs or drawbacks, providing a reference goal that directs designers toward solutions beyond conventional design boundaries.
\end{itemize}

\subsubsection{Contradiction matrix}
\label{subsubsec:contradiction_matrix}
\begin{figure}[h]
\centering
\centering\includegraphics[width=1.00\linewidth]{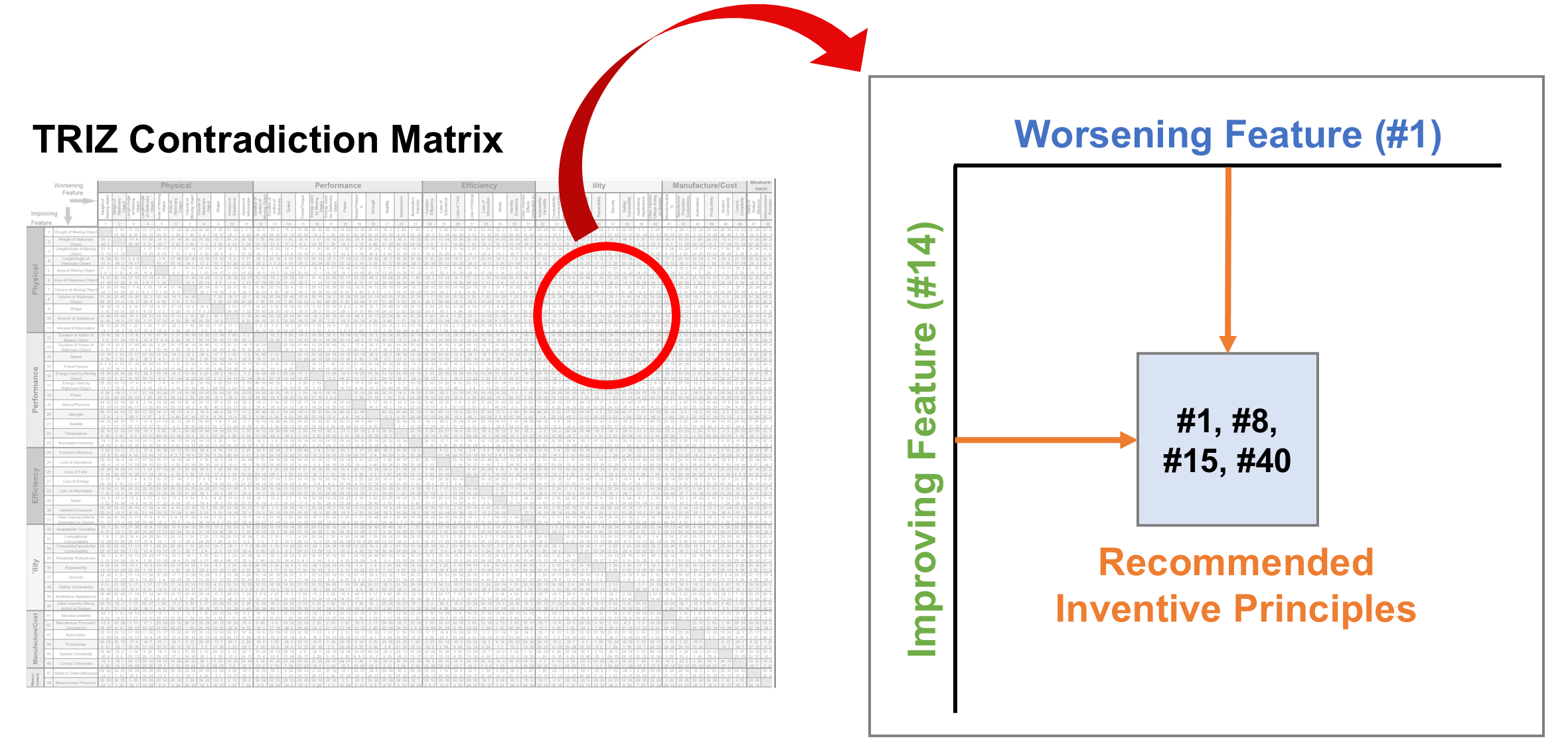}
\caption{Illustrative example of the TRIZ contradiction matrix (strength (\#14) vs. weight (\#1)).}
\label{fig:contradiction_matrix}
\end{figure}

The contradiction matrix is a practical tool that cross-references 39 improving features with 39 worsening features, recommending specific inventive principles for each combination, as illustrated in Fig.~\ref{fig:contradiction_matrix} \cite{triz2003matrix, mann2003updating}. For example, when attempting to improve \textit{Strength} (Feature \#14) while avoiding increase in \textit{Weight of moving object} (Feature \#1), the matrix recommends inventive principles such as \textit{Segmentation} (\#1), \textit{Anti-weight} (\#8), \textit{Dynamics} (\#15), and \textit{Composite Materials} (\#40). This systematic approach to innovation makes TRIZ particularly suitable for integration with LLMs, as the inventive principles and contradiction matrix can be encoded as contextual guidance within prompts to effectively support creative design exploration.

\subsubsection{TRIZ applications with LLMs}
Recent studies have explored generative AI for TRIZ reasoning, including TRIZ-GPT \cite{chen2024triz}, TRIZ-inspired eco-design \cite{lee2024generating}, and AutoTRIZ \cite{jiang2025autotriz}, demonstrating the feasibility of combining TRIZ with LLMs at a conceptual design level. However, these integrations remain confined to ideation stages and do not connect TRIZ-based reasoning to executable CAD code, motivating the present work to embed TRIZ-informed reasoning directly into the text-to-CAD generation loop to enable creative design exploration.


\section{Methodology: TRIZ-Inspired Text-to-CAD Framework for Creative 3D Model Generation}
\label{sec:methodology}

\begin{figure*}[h!]
\centering
\includegraphics[width=1.00\linewidth]{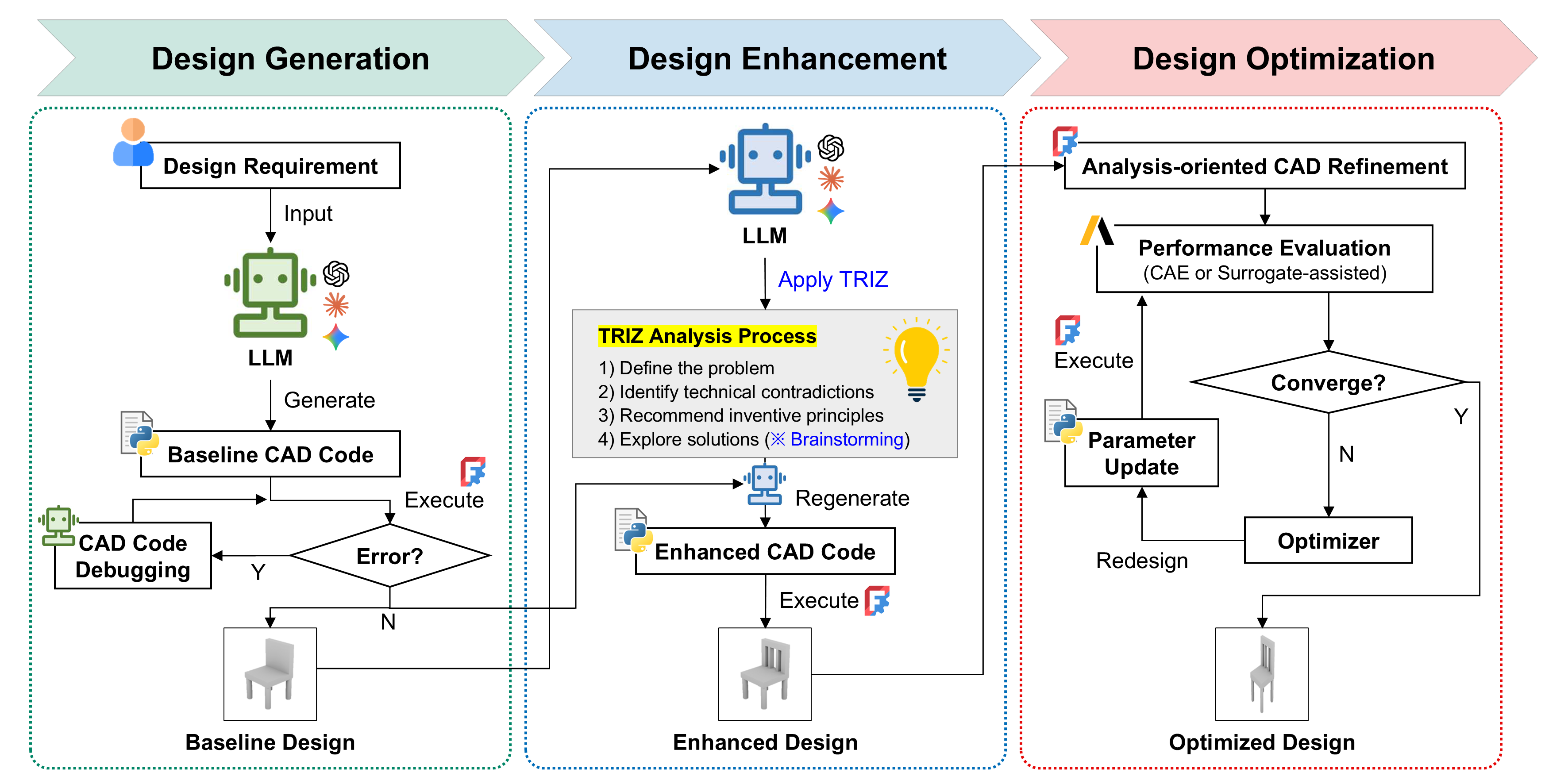}
\caption{Overview of the proposed TRIZ-inspired text-to-CAD framework. The framework progresses from (1) design generation through domain-specific prompting, (2) design enhancement via TRIZ inventive principle application, to (3) design optimization using computational analysis for engineering performance refinement.}
\label{fig:research_framework}
\end{figure*}

\subsection{Framework Overview}
\label{subsec:framework_overview}
Fig.~\ref{fig:research_framework} illustrates the proposed TRIZ-inspired Text-to-CAD framework consisting of three stages:

\begin{itemize}
    \item \textbf{Design generation}: Baseline parametric CAD models are generated from natural language requirements via domain-specific prompt engineering, producing executable CAD code with well-defined parameters and geometric validity checks. This stage can be omitted when an editable baseline model is already available.
    \item \textbf{Design enhancement}: Creative design alternatives are systematically produced by identifying technical contradictions, retrieving relevant inventive principles from the contradiction matrix, and generating enhanced designs that resolve these contradictions while maintaining parametric editability and geometric validity.
    \item \textbf{Design optimization}: Design parameters are iteratively refined for engineering performance using computational analysis such as CAE simulation, leveraging the parametric structure established in previous stages to enable automated design optimization.
\end{itemize}

This study focuses on Stages 1 and 2, which constitute the core methodological contribution. Rather than treating TRIZ as a post-hoc ideation tool, the framework embeds TRIZ analysis—including inventive principles and contradiction matrix—directly into LLM prompts, enabling systematic exploration of creative design solutions while maintaining parametric editability and geometric validity. The framework adopts a structured prompt format consisting of four elements: \textit{Role}, \textit{Task}, \textit{Requirements}, and \textit{Expected Output}, with the TRIZ-informed prompt extending this baseline with contextual guidance to resolve design contradictions—transforming LLMs from passive code generators into autonomous design agents capable of reasoning about design trade-offs through knowledge-guided prompts.

\subsection{Stage 1: Design generation}
\label{subsec:design_generation}
\begin{figure*}[h!]
\centering
\includegraphics[width=1.00\linewidth]{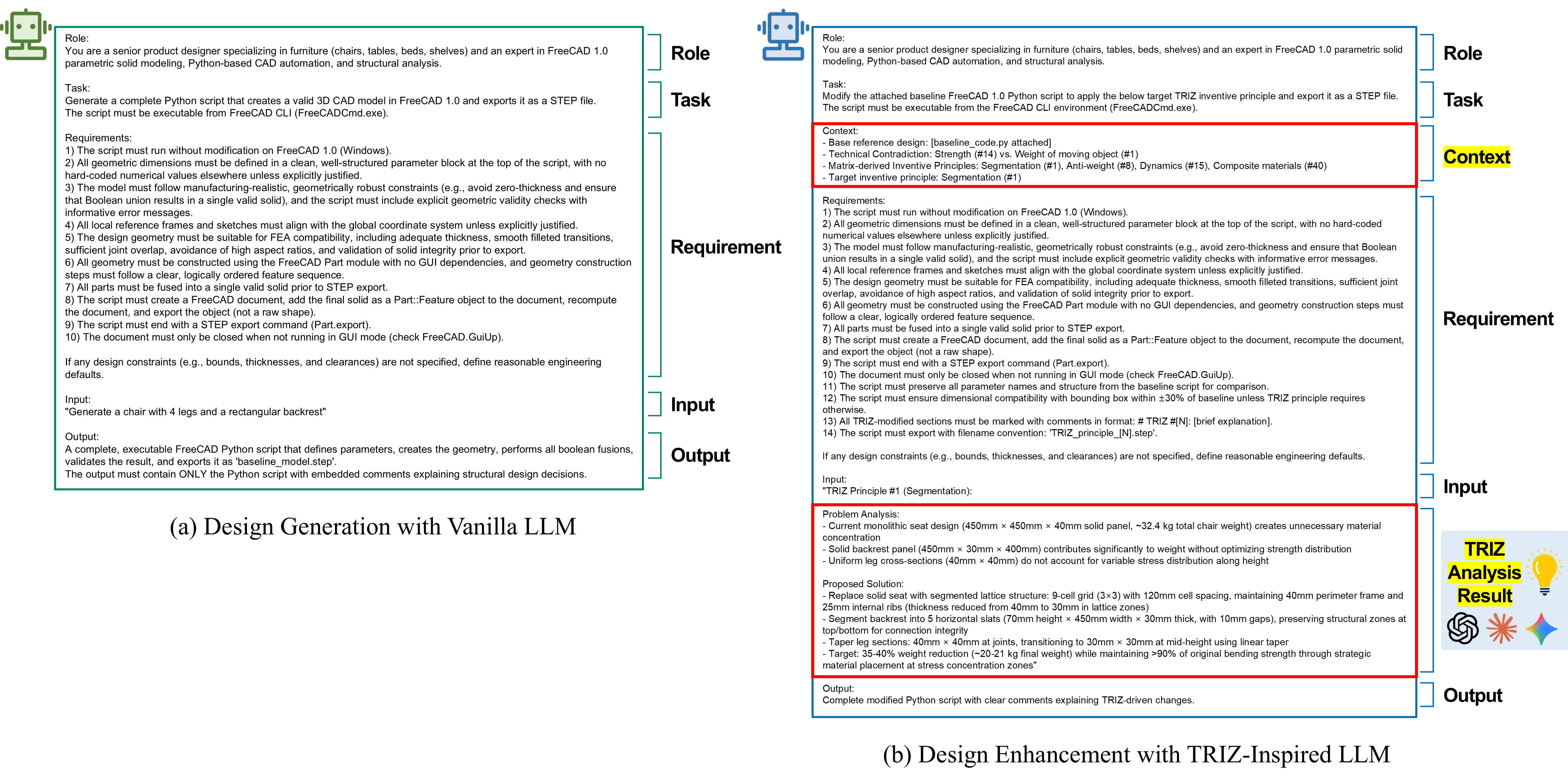}
\caption{Examples of structured prompts for (a) baseline design generation and (b) TRIZ-inspired design enhancement. Red boxes highlight TRIZ-specific additions including recommended principles and structured problem analysis applied to a chair design case.}
\label{fig:prompt_examples}
\end{figure*}

The design generation stage produces baseline parametric CAD models from natural language prompts. As shown in Fig.~\ref{fig:prompt_examples}(a), the baseline prompt specifies: (1) the LLM's role as a CAD expert, (2) the task of generating executable Python scripts for FreeCAD, (3) detailed requirements including platform compatibility, parametric design practices, geometric validity checks, and proper export procedures, and (4) the input and expected output. The generated code is immediately executed with iterative debugging if errors occur. The prompt prioritizes parametric practices—variable-based dimensions, comprehensive comments, and modular code structure—to enable TRIZ-based modifications in Stage 2, ensuring controllability and reproducibility.

\subsection{Stage 2: Design enhancement}
\label{subsec:design_enhancement}
Stage 2 integrates TRIZ methodology directly into the prompt structure to enable systematic creative design exploration. The process begins with TRIZ-based problem analysis to identify technical contradictions, followed by retrieval of relevant inventive principles from the TRIZ contradiction matrix. As shown in Fig.~\ref{fig:prompt_examples}(b), the TRIZ-informed prompt extends the baseline format with: (1) a context section specifying the baseline reference, identified contradiction, and recommended TRIZ principles, and (2) the results of problem analysis and the proposed solutions in the TRIZ analysis results section comprising problem identification and proposed solutions. The LLM interprets these principles, generating parametrically editable and geometrically valid CAD code that resolves the identified contradictions, with multiple alternatives produced by varying principle combinations.

\subsection{Stage 3: Design optimization}
\label{subsec:design_optimization}
As noted above, this study presents a three-stage framework of which Stages 1 and 2 are implemented and empirically evaluated. While the framework conceptually integrates a third stage for performance-driven optimization through CAE simulation and surrogate modeling, this study focuses on Stages 1 and 2 to evaluate structured prompt engineering for creative CAD exploration. Stage 3 would involve iterative parameter refinement using optimization algorithms, with potential LLM participation in interpreting simulation results and suggesting adjustments. Importantly, the feasibility of Stage 3 depends critically on the parametric code quality established in Stages 1 and 2. The complete integration of Stage 3, including LLM-assisted optimization strategies, represents future work.


\section{Results and discussion}
\label{sec:results_and_discussion}

\subsection{Experimental setup}
\label{subsec:experimental_setup}
All experiments were conducted using a closed-source LLM for CAD code generation. We used Claude Sonnet 4.5 via a web-based interface, without API-level control, model fine-tuning, or additional system customization. A text-only prompt modality was employed, in which natural language design descriptions were combined with well-crafted TRIZ-informed contextual guidance. Prompt engineering was used to explicitly encode design intent, parametric constraints, and TRIZ-based inventive principles, enabling systematic and controlled creative design exploration without modifying the underlying model. To mitigate the non-deterministic nature of LLM inference, the prompt format and wording were iteratively refined through trial-and-error to encourage as consistent and reproducible outputs as possible. All generated CAD code was executed and validated in a standard CAD environment (FreeCAD 1.0) based on the following criteria: (1) successful script execution, (2) preservation of parametric editability, and (3) geometric validity of the resulting solid model and STEP exportability. No post-processing, external retrieval mechanisms, or model fine-tuning were applied, in order to evaluate the feasibility of the proposed framework using widely available closed-source LLMs.

\subsection{Case study description}
\label{subsec:case_study_description}
A simple chair design was selected as a demonstrative case study due to its structural simplicity, clearly defined functional requirements, and suitability for illustrating representative strength–weight trade-offs in early-stage conceptual design. This case study enables the effects of the TRIZ-inspired text-to-CAD framework on geometric variation and structural characteristics to be examined in a systematic, controlled and interpretable setting. Following a performance-focused design objective, a technical contradiction was formulated between improving structural strength (improving feature \#14) and minimizing weight (worsening feature \#1), as shown in Fig.~\ref{fig:contradiction_matrix}. The TRIZ contradiction matrix identified the corresponding inventive principles—segmentation (\#1), anti-weight (\#8), dynamics (\#15), and composite materials (\#40)—which were embedded as structured contextual guidance within the design enhancement prompts.

The TRIZ problem formulation is as follows:

\begin{itemize}
    \item \textbf{Design objective}: Performance-focused approach considering strength and weight trade-offs
    \item \textbf{Improving feature}: Strength (\#14)
    \item \textbf{Worsening feature}: Weight of moving object (\#1)
    \item \textbf{Recommended inventive principles}:
    \begin{enumerate}
        \item Segmentation (\#1)
        \item Anti-weight (\#8)
        \item Dynamics (\#15)
        \item Composite Materials (\#40)
    \end{enumerate}
\end{itemize}


\clearpage
\subsection{Design generation (Baseline)}
\label{subsec:design_generation_results}

\subsubsection{Generated CAD code and 3D model}
\label{subsubsec:generated_code_and_model}
\begin{figure}[h]
\centering
\includegraphics[width=1.00\linewidth]{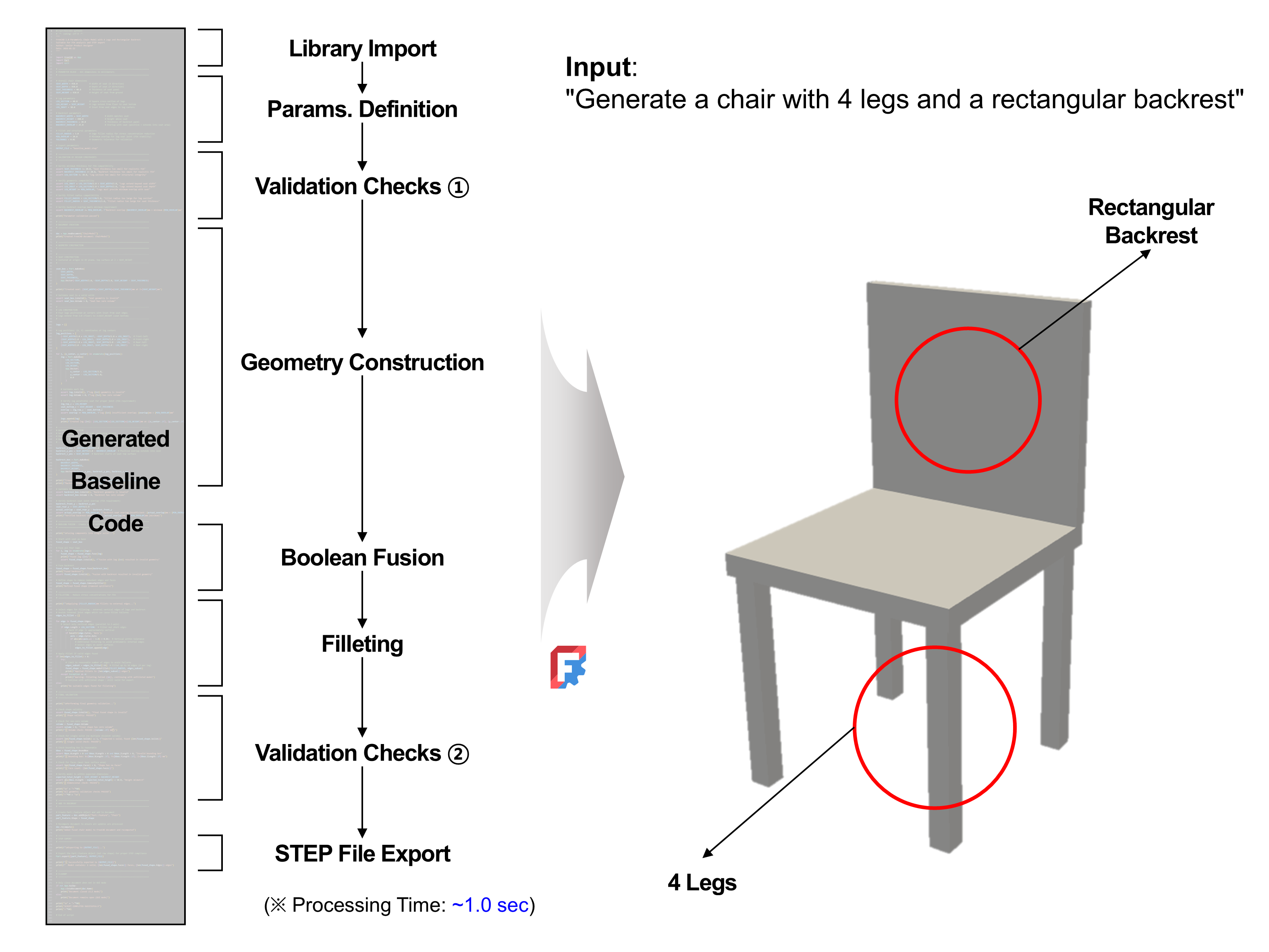}
\caption{Overview of the generated baseline CAD code, its generation workflow, and the resulting 3D chair model.}
\label{fig:generated_code_model}
\end{figure}

The baseline chair design was generated from the natural language input: \textit{"Generate a chair with 4 legs and a rectangular backrest"}. Fig.~\ref{fig:generated_code_model} illustrates both the functional workflow of the generated Python script and the resulting 3D model. The LLM successfully produced a complete CAD generation script comprising 302 lines of well-structured code that executes a systematic workflow: library import, parametric definition, validation checks, geometry construction, Boolean fusion, filleting, and STEP file export.

The script demonstrates structured parametric CAD modeling practices through hierarchical parameter organization, explicit validation checks, and modular geometry construction, as summarized in Table~\ref{tab:baseline_code_details}.

\begin{table}[h]
\centering
\caption{Generated baseline CAD code details}
\label{tab:baseline_code_details}
\small
\begin{tabularx}{\columnwidth}{ll>{\raggedleft\arraybackslash}X}
\hline
\multicolumn{1}{c}{\textbf{Metric}} & \multicolumn{1}{c}{\textbf{Description}} & \multicolumn{1}{c}{\textbf{Value}} \\
\hline
Total lines of code & Full script & 302 \\
Lines of comments & Inline documentation & 98 \\
Parameter definitions & Design variables & 14 \\
Validation assertions & Constraint checks & 25 \\
Geometric components & Seat, Legs (4), Backrest & 6 \\
Boolean operations & Sequential fusions & 5 \\
Execution time (sec) & Baseline run & $<$1.0 \\
\hline
\end{tabularx}
\end{table}

Notably, 14 parametric variables control all geometric dimensions, organized into logical groups encompassing seat, leg, backrest, and auxiliary parameters, facilitating efficient design iteration. The script also incorporates a three-stage validation strategy with 25 assertions covering pre-construction constraints, component-level integrity checks, and post-fusion model validity, collectively reducing downstream Boolean failures and ensuring geometric reliability. The geometry construction follows a modular workflow in which individual components—seat, four legs, and backrest—are created as box solids using \texttt{Part.makeBox()}, with deliberate overlaps introduced at structural joints to ensure proper connectivity and FEA meshing stability. These components are sequentially merged through Boolean fusion operations into a single solid body, followed by \texttt{removeSplitter()} refinement and a filleting stage with exception handling to maintain robustness even when filleting fails. The script executed successfully in FreeCAD 1.0 within 1 second, generating a valid STEP file, which confirms that the LLM can produce not only syntactically valid code but also functionally consistent and parametrically organized CAD code—making this approach substantially more practical than non-parametric shape generation for manufacturing-oriented design workflows.

\subsubsection{Baseline design validation and specifications}
\label{subsubsec:baseline_validation}
The generated baseline design successfully satisfied all validation criteria established in Section~\ref{subsec:experimental_setup}. Table~\ref{tab:baseline_validation} summarizes the validation results across 5 critical aspects: script execution, parametric constraints, component joints, geometric validity, and STEP export compatibility. The Python script executed without errors in FreeCAD 1.0 within 1 second, producing a geometrically valid single solid body and satisfying all 25 validation assertions related to parametric constraints and component joints. Moreover, the model was successfully exported to STEP format with proper model structure, confirming downstream CAE compatibility.

\begin{table}[h]
\centering
\caption{Baseline design validation results}
\label{tab:baseline_validation}
\small
\begin{tabularx}{\columnwidth}{Xc}
\hline
\multicolumn{1}{c}{\textbf{Validation Criterion}} & \multicolumn{1}{c}{\textbf{Status}} \\
\hline
Script execution (FreeCAD 1.0) & \textcolor{mydarkgreen}{\CheckmarkBold} \\
Parametric constraints (length, thickness, section) & \textcolor{mydarkgreen}{\CheckmarkBold} \\
Component joints (overlap) & \textcolor{mydarkgreen}{\CheckmarkBold} \\
Geometric validity (single solid, positive volume) & \textcolor{mydarkgreen}{\CheckmarkBold} \\
STEP export compatibility & \textcolor{mydarkgreen}{\CheckmarkBold} \\
\hline
\end{tabularx}
\end{table}

Table~\ref{tab:baseline_specs} provides the detailed dimensional specifications of the baseline model (a simple chair). As summarized in the table, all dimensional constraints are satisfied, with joint overlaps at both the legs-seat and backrest-seat interfaces meeting finite element analysis (FEA) meshing requirements. The resulting baseline represents a structurally conservative design with substantial potential for mass reduction, while providing the parametric flexibility necessary for subsequent TRIZ-based design enhancement explored in Section~\ref{subsec:design_enhancement_results}.

\begin{table}[h]
\centering
\caption{Baseline design specifications}
\label{tab:baseline_specs}
\small
\begin{tabularx}{\columnwidth}{ll>{\raggedleft\arraybackslash}X}
\hline
\multicolumn{1}{c}{\textbf{Category}} & \multicolumn{1}{c}{\textbf{Parameter}} & \multicolumn{1}{c}{\textbf{Value}} \\
\hline
\multirow{3}{*}{Overall} 
& Bounding box (L$\times$W$\times$H) (mm) & 455$\times$450$\times$850 \\
& Volume (m$^3$) & 1.61$\times$10$^{-2}$ \\
& Mass\textsuperscript{*} (kg) & 16.93 \\
\hline
\multirow{5}{*}{Components} 
& Seat (W$\times$D$\times$T) (mm) & 450$\times$450$\times$40 \\
& Backrest (W$\times$H$\times$T) (mm) & 450$\times$400$\times$30 \\
& Leg, cross-section (4 ea.) (mm) & 40$\times$40 \\
& Leg, height (mm) & 450 \\
& Leg, inset from seat edge (mm) & 30 \\
\hline
\multirow{2}{*}{Joints} 
& Legs-seat overlap (mm) & 20 \\
& Backrest-seat overlap (mm) & 25 \\
\hline
\multicolumn{3}{@{}p{\dimexpr\columnwidth-2\tabcolsep}@{}}{\footnotesize\textsuperscript{*}Mass was calculated assuming ABS plastic with a density of $\rho = 1050$~kg/m$^3$.}
\end{tabularx}
\end{table}


\subsection{Design enhancement (TRIZ-inspired alternatives)}
\label{subsec:design_enhancement_results}

\subsubsection{Finite element analysis setup}
\label{subsubsec:fea_setup}
To assess the structural implications of the generated designs, static FEA was conducted using ANSYS Mechanical 2023 R2. All models were assigned ABS plastic material properties with fixed support constraints at the four leg bases and a 900 N vertical load on the seat surface, representing an average adult male (approximately 75 kg, ~750 N) with a 20\% dynamic safety margin. Standard gravity (9.81 m/s$^2$) was also applied for structural self-weight. Automatic meshing with default settings provided adequate accuracy for comparative evaluation. Three key metrics were extracted: mass, equivalent von Mises stress, and total deformation. All boundary conditions and loads were defined consistently across variants to enable fair comparative assessment of strength (\#14)--weight (\#1) trade-offs, with FEA results supporting structural behavior discussion rather than absolute design validation.

\begin{figure}[h]
\centering
\includegraphics[width=1.00\linewidth]{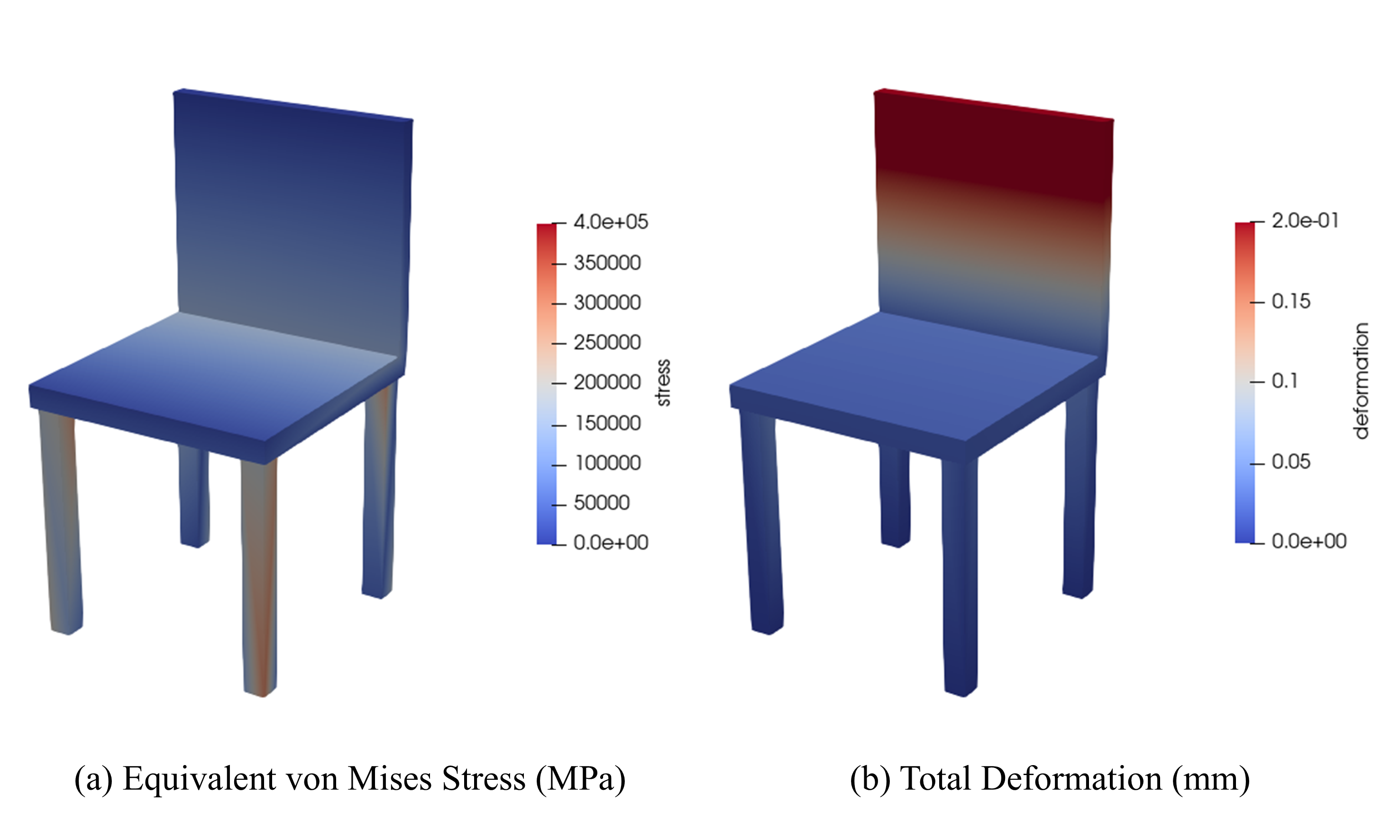}
\caption{Structural analysis results of baseline design showing (a) equivalent von Mises stress and (b) total deformation distribution under static loading (900 N vertical force on seat surface).}
\label{fig:fea_results}
\end{figure}

Prior to comparative analysis of TRIZ-inspired design alternatives, the baseline design was analyzed to verify the FEA setup appropriateness. Fig.~\ref{fig:fea_results} shows the structural response of the baseline simple chair under 900 N loading, with maximum von Mises stress of 0.627 MPa (1.7\% of material yield strength) and maximum total deformation of 0.288 mm at the seat center. These results confirm extremely conservative structural margins in the baseline design, indicating substantial potential for mass reduction through TRIZ-inspired design enhancement while maintaining adequate safety factors.

\subsubsection{TRIZ-inspired design alternatives and performance comparison}
\label{subsubsec:design_alternatives}
\begin{figure*}[h!]
\centering
\includegraphics[width=1.00\linewidth]{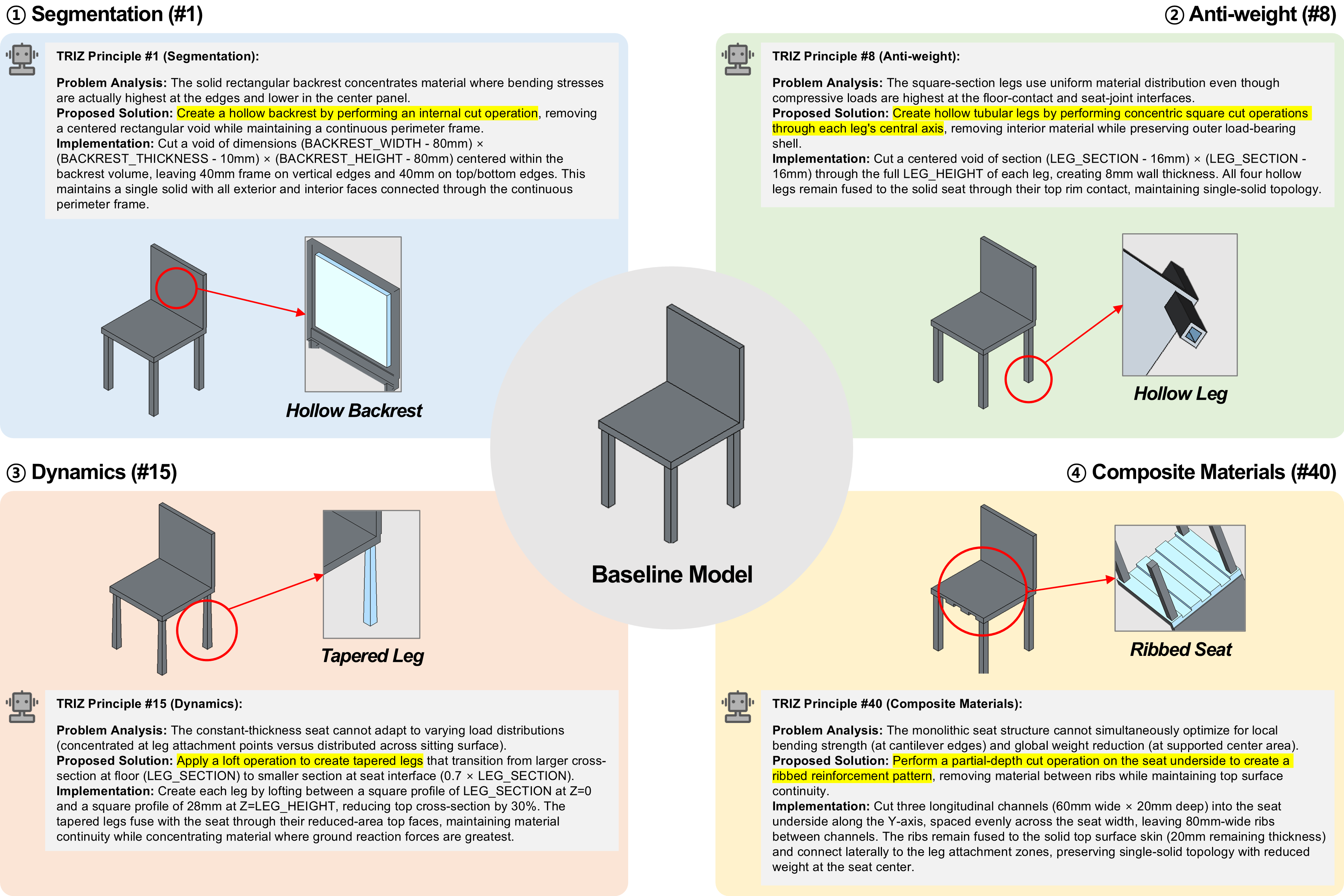}
\caption{TRIZ-inspired design alternatives addressing strength-weight contradiction: (1) hollow backrest via segmentation, (2) hollow leg via anti-weight, (3) tapered leg via dynamics, and (4) ribbed seat via composite materials.}
\label{fig:triz_alternatives}
\end{figure*}

Fig.~\ref{fig:triz_alternatives} and Table~\ref{tab:fea_comparison} present the TRIZ-inspired design alternatives and their comparative structural performance, each applying a specific inventive principle to address the strength (\#14) vs. weight (\#1) contradiction. The segmentation (\#1) principle produced a hollow backrest design by removing internal material while preserving the outer frame, achieving 14.7\% mass reduction with negligible stress increase (-0.1\%). The anti-weight (\#8) principle introduced hollow tubular legs through concentric square-cut operations, yielding 5.9\% mass reduction with moderate stress increase (+21.6\%). The dynamics (\#15) principle applied lofting operations to create tapered legs with reduced cross-sections at mid-height, resulting in 4.0\% mass reduction but 33.2\% stress increase due to reduced cross-sectional area. The composite materials (\#40) principle implemented a ribbed seat pattern with longitudinal channels, achieving 10.0\% mass reduction but the highest stress increase (+40.7\%) from stress concentration at rib edges. Each design modification was implemented through basic parametric operations (extrude-cut, loft, pattern) compatible with conventional CAD workflows, ensuring manufacturability and parametric editability for subsequent design iterations.

\begin{table}[h]
\centering
\caption{Comparative structural performance of TRIZ-inspired design alternatives}
\label{tab:fea_comparison}
\small
\begin{tabularx}{\columnwidth}{Xrrr}
\hline
\multicolumn{1}{c}{\textbf{Design alternative}} & 
\multicolumn{1}{c}{\makecell{\textbf{Mass} \\ (kg)}} & 
\multicolumn{1}{c}{\makecell{\textbf{Max stress} \\ (MPa)}} & 
\multicolumn{1}{c}{\makecell{\textbf{Max disp.} \\ (mm)}} \\
\hline
Baseline & \makecell[r]{16.930} & \makecell[r]{0.627} & \makecell[r]{0.288} \\
\hline
Vanilla Prompt & \makecell[r]{11.552 \\ (-31.8\%)} & \makecell[r]{1.028 \\ (+64.1\%)} & \makecell[r]{0.574 \\ (+99.4\%)} \\
\hline
\makecell[l]{TRIZ \#1 \\ (Segmentation)} & \makecell[r]{14.444 \\ (-14.7\%)} & \makecell[r]{0.626 \\ (-0.1\%)} & \makecell[r]{0.290 \\ (+0.7\%)} \\
\makecell[l]{TRIZ \#8 \\ (Anti-weight)} & \makecell[r]{15.938 \\ (-5.9\%)} & \makecell[r]{0.762 \\ (+21.6\%)} & \makecell[r]{0.301 \\ (+4.4\%)} \\
\makecell[l]{TRIZ \#15 \\ (Dynamics)} & \makecell[r]{16.246 \\ (-4.0\%)} & \makecell[r]{0.835 \\ (+33.2\%)} & \makecell[r]{0.335 \\ (+16.3\%)} \\
\makecell[l]{TRIZ \#40 \\ (Composite Materials)} & \makecell[r]{15.229 \\ (-10.0\%)} & \makecell[r]{0.882 \\ (+40.7\%)} & \makecell[r]{0.416 \\ (+44.4\%)} \\
\hline
\end{tabularx}
\end{table}

All designs successfully reduced mass (4.0-14.7\%) while maintaining structural integrity, with maximum stresses of 0.626-0.882 MPa—well below the ABS plastic yield strength. The stress increases reflect the fundamental lightweight design trade-off: strategic material removal concentrates stresses in remaining load-bearing regions. However, these elevated stresses remain within acceptable engineering margins, validating the viability of TRIZ-inspired design modifications for practical implementation. Similarly, maximum displacements increased modestly from 0.288 mm to 0.290-0.416 mm, with the segmentation alternative showing minimal change (+0.7\%) and composite materials the largest increase (+44.4\%). Nonetheless, all deformation magnitudes remain below 0.5 mm—negligible relative to the 850 mm overall chair height and well within acceptable deflection limits for furniture applications.

These results demonstrate that the proposed TRIZ-inspired text-to-CAD framework successfully resolves the strength-weight contradiction by generating diverse and structurally valid design alternatives that achieve meaningful mass reduction (averaging 8.7\%) while preserving structural performance within safe operating margins. This balance represents a key advantage for early-stage design innovation where rapid iteration and broad concept generation across a wide design space are valued over exhaustive numerical optimization.

To isolate the contribution of TRIZ-informed prompting, a vanilla prompt baseline was evaluated using the instruction: \textit{"Modify the attached baseline FreeCAD Python script to create a lighter version of the chair"}. As shown in Table~\ref{tab:fea_comparison}, the vanilla prompt achieved a 31.8\% mass reduction by scaling down geometric dimensions without structural modifications. However, this quantitative scaling strategy led to a 64.1\% stress increase and a 99.4\% displacement increase, substantially exceeding the trade-offs observed across the TRIZ-inspired design alternatives, where stress changes ranged from -0.1\% to +40.7\% and displacement increases remained within +44.4\%. This contrast highlights a key difference, as the vanilla prompt drives the LLM toward quantitative scaling without structural reasoning, whereas TRIZ-informed prompting supports qualitative transformation by explicitly encoding the strength-weight contradiction.

\subsubsection{Readiness for design optimization}
\label{subsubsec:readiness_optimization}

Beyond structural performance, the generated design alternatives collectively demonstrate readiness for design optimization as envisioned in Stage 3 of the proposed framework. Each variant is defined by well-structured parametric variables inherited from the baseline model, enabling gradient-based or gradient-free optimization over key dimensions, and the FEA-validated responses confirm simulation compatibility for subsequent optimization through direct CAE simulation or surrogate-assisted approaches. Taken together, the parametric editability, geometric validity, and CAE compatibility of the TRIZ-inspired design alternatives establish a sufficiently robust foundation upon which the design optimization stage can be seamlessly integrated in future work.


\subsection{Limitations}
\label{subsec:limitations_and_practical_implications}
This study acknowledges the inherent non-deterministic nature of LLM-based code generation. Due to probabilistic token sampling mechanisms controlled by temperature, top-p, and top-k parameters, LLMs may produce syntactically different code implementations for identical inputs, even when using structured prompts. In our experiments, this variability primarily affected stylistic aspects—such as variable naming, code organization, and intermediate calculations—rather than functional outcomes. All generated models maintained geometric correctness and satisfied parametric design specifications, demonstrating that while exact code reproducibility cannot be guaranteed, functional reproducibility—generating geometrically equivalent and manufacturable CAD models—was reliably achieved.

In addition, text-based design requirements in Stage 1 may initially lead to CAD generation errors, including invalid Boolean operations or geometric inconsistencies, which were addressed through iterative debugging within the proposed framework (Fig.~\ref{fig:research_framework}), where failure cases informed subsequent prompt adjustments. Although this debugging process was performed iteratively through interactive LLM-assisted refinement, the identified error patterns and corrective actions highlight an important limitation of current text-to-CAD workflows and point toward future extensions of the framework, in which debugging knowledge and validation rules can be systematically incorporated to improve robustness. Furthermore, TRIZ-inspired design alternatives in Stage 2 occasionally produced conceptually innovative solutions that exceeded the geometric complexity of the baseline model or required CAD operations beyond basic parametric features, limiting their direct implementation feasibility. In such cases, simplification or adaptation was necessary to maintain CAD compatibility with standard parametric modeling capabilities.


\section{Conclusions and future work}
\label{sec:conclusions_and_future_work}
This study presented a TRIZ-inspired text-to-CAD framework for integrating structured inventive principles into LLM-based CAD code generation via TRIZ-informed prompts. By embedding TRIZ context directly into prompts for widely available closed-source LLMs, the proposed approach enables systematic exploration of design alternatives while maintaining parametric editability and geometric validity. A representative chair design case study demonstrated that the framework generates four structurally diverse CAD alternatives, achieving 4.0-14.7\% mass reduction (averaging 8.7\%) while maintaining stresses well within the material yield strength, validating knowledge-guided prompt engineering as an effective bridge between precision-focused CAD generation and creative design exploration. As generative AI continues to reshape engineering design practices, this work provides foundational insights into how structured knowledge frameworks can guide LLMs toward more systematic, creative, and autonomous design generation.

Although grounded in prompt engineering—an accessible entry point for integrating TRIZ context into LLMs—the underlying methodology is not confined to this paradigm. As the AI landscape evolves through successive paradigms, from prompt engineering toward context engineering, harness engineering, and governance engineering, and beyond, the proposed framework is expected to remain applicable and progressively more effective; the core principle of embedding structured innovation knowledge into AI-driven design workflows remains valid regardless of which interaction mechanism is adopted.

Several directions for future work are identified. First, systematic prompt optimization under API-based controlled settings for reproducibility—including edit-based, generation-based, text gradient-based, and evolutionary approaches—will be explored to further refine TRIZ-informed prompts and enhance design space exploration, treating prompts as optimizable components. Second, support for multi-modal inputs (e.g., sketches, images) and complex design operations (e.g., parametric constraints, assemblies) would broaden the framework's applicability across diverse design contexts. Furthermore, incorporating human evaluation—whereby designers or engineers rate generated alternatives on novelty, usefulness, and feasibility—would establish a human-on-the-loop feedback mechanism, providing complementary qualitative validation and further guiding the refinement of the framework toward higher-quality creative design outcomes. Third, the framework can be extended to integrate automated CAD/CAE workflows and complete the design optimization stage through surrogate modeling and optimization algorithms, thereby realizing a fully end-to-end autonomous design pipeline. Fourth, extending the methodology to diverse engineering domains could enhance its generalizability beyond structural product design. Fifth, future efforts will focus on the integration of ontology-inspired structured knowledge frameworks within generative AI-driven engineering workflows to improve controllability, scalability, and systematic design exploration.


\section*{Acknowledgments}
This work was supported by grants from the Ministry of Science and ICT (GTL24031-000, N10250154, No.2022-0-00986, and RS-2024-00355857) and the Ministry of Trade, Industry \& Energy (RS-2025-02317327, RS-2025-25444634).


\nocite{*}

\bibliographystyle{asmeconf}  
\bibliography{references}






\end{document}